\documentclass{article}
\PassOptionsToPackage{numbers, compress}{natbib}
\usepackage[final]{neurips_2019}

\usepackage{amssymb}
\usepackage{hyperref}
\usepackage{times}  % DO NOT CHANGE THIS
\usepackage{helvet}  % DO NOT CHANGE THIS
\usepackage{courier}  % DO NOT CHANGE THIS
\usepackage{graphicx} % DO NOT CHANGE THIS
\usepackage{natbib}  % DO NOT CHANGE THIS AND DO NOT ADD ANY OPTIONS TO IT
\usepackage{caption} % DO NOT CHANGE THIS AND DO NOT ADD ANY OPTIONS TO IT

\usepackage{algorithm}
\usepackage{algpseudocode}

\usepackage[table,xcdraw]{xcolor}
\usepackage[utf8]{inputenc} % allow utf-8 input
\usepackage[T1]{fontenc}    % use 8-bit T1 fonts
\usepackage{url}            % simple URL typesetting
\usepackage{booktabs}       % professional-quality tables
\usepackage{amsfonts}       % blackboard math symbols
\usepackage{amsmath}
\usepackage{amsthm}
\usepackage{nicefrac}       % compact symbols for 1/2, etc.
\usepackage{xcolor}         % colors

\usepackage{xspace}
\usepackage{multirow}
\usepackage{makecell}
\usepackage{array}
\usepackage{subcaption}
\usepackage{kotex}
\usepackage{enumitem}
\usepackage{comment}
\usepackage{pifont}
\usepackage{wrapfig}
\usepackage[font=small]{caption}
\usepackage{hyperref}
\usepackage{booktabs}
\usepackage{bbm}
\usepackage{adjustbox}

\title{Invisible Watermarks: Attacks and Robustness}

\author{
Dongjun Hwang$^{1}$\thanks{Equal contribution.} \qquad Sungwon Woo$^{1*}$ \qquad Tom Gao$^{1*}$ \\[0mm]
\textbf{Raymond Luo}$^{1*}$ \qquad \textbf{Sunghwan Baek}$^{1}$ \vspace{.5em} \\[0mm]
Project Group 3~\thanks{
This paper serves as the final project report for the Introduction to Deep Learning (11-785) course at Carnegie Mellon University.}\\
$^1$Carnegie Mellon University \\ Pittsburgh, PA 15213\\
\texttt{\{dongjunh,sungwonw,zimingg,rluo2,sunghwab\}@andrew.cmu.edu} \\
}

\newcommand{\myparagraph}[1]{\vspace{0pt}\noindent{\bf #1}}

\begin{document}

\maketitle

% You can refer the like this~\cite{PAPER}. If you want to cite the paper, then you should add bibtex format to \texttt{ref.bib}.

% You can find the bibtext format of the paper using the google scholar -> search paper -> cite -> BibTex.

\begin{abstract}
As Generative AI continues to become more accessible, the case for robust detection of generated images in order to combat misinformation is stronger than ever. Invisible watermarking methods act as identifiers of generated content, embedding image- and latent-space messages that are robust to many forms of perturbations. The majority of current research investigates full-image attacks against images with a single watermarking method applied. We introduce novel improvements to watermarking robustness as well as minimizing degradation on image quality during attack. Firstly, we examine the application of both image-space and latent-space watermarking methods on a single image, where we propose a custom watermark remover network which preserves one of the watermarking modalities while completely removing the other during decoding. Then, we investigate localized blurring attacks (LBA) on watermarked images based on the GradCAM heatmap acquired from the watermark decoder in order to reduce the amount of degradation to the target image. Our evaluation suggests that 1) implementing the watermark remover model to preserve one of the watermark modalities when decoding the other modality slightly improves on the baseline performance, and that 2) LBA degrades the image significantly less compared to uniform blurring of the entire image. Code is available at: \url{https://github.com/tomputer-g/IDL_WAR}
\end{abstract}

\section{Introduction}

\myparagraph{Overview of Watermarking.} Watermarks serve as digital markers to indicate ownership or embed a message. A user can use an input image, apply a watermark message or key   , and then use a decoder to extract and interpret the watermark. There are two main types of watermarks currently: \textbf{visible watermarks} and \textbf{invisible watermarks}. Visible watermarks, such as logos or text overlays, are simple to create but often degrade image aesthetics and are easily removed using advanced tools like Stable Diffusion~\citep{stable-diffusion}. In contrast, invisible watermarks embed markers directly into an image, preserving its visual quality. This shift toward invisible watermarking has spurred the development of robust embedding techniques and corresponding attack methods, especially in the context of machine learning-based approaches. With the rise of image generating latent diffusion models~\citep{stable-diffusion}, watermarking robustness is more important than ever to ensure that we can differentiate between what images are real and what images are AI-generated.

\myparagraph{Our Contribution.} We investigate improving watermark robustness by combining machine learning-based watermarking techniques and separately improving attacks against a particular watermark known as StegaStamp by specifically attacking pixels important to the watermark. Through these contributions, we hope to improve watermark robustness by building in redundancy between the multiple watermarks and improve image quality degradation from watermark attacks by reducing the number of pixels that we attack.

\section{Literature Review}
\label{sec:literature_review}
\subsection{Watermarking Methods}
Currently, there are four popular AI based invisible image watermarking methods: StegaStamp, Tree-Ring, Stable Signature, and Latent Watermark. In this paper, we focus on StegaStamp and Tree-Ring for implementation, and use Stable Signature watermarked images as part of our baseline evaluation.

StegaStamp is a CNN-based watermarking method using a learned steganographic algorithm, which encodes the given message into the generated image as a post-processing step  \citep{stegastamp}. Because the encoder and decoder network are trained against common image perturbations, the watermark is fairly robust against common attacks.

Tree-Ring modifies the distribution of generated images by embedding a watermark into the initial noise vector of diffusion models, which can be algorithmically retrieved at detection time \citep{tree-ring}. This creates a watermark that is relatively robust against some common image manipulation attacks, but it greatly changes how the image looks (only maintaining semantic meaning).

Stable Signature is a method to fine-tune the latent decoder of the generative model, which is conditioned to embed each generated image with the model's unique signature \citep{stable-signature}. During the detection step, the pre-trained watermark extractor recovers the signature and performs a statistical test to detect the origin of the image.

\subsection{Watermark Attack Methods}

Current watermark attacking methods loosely fall into the following categories: distortion, regeneration, and adversarial~\citep{WAVES}. 

Regeneration attacks will be our main baseline for this paper. Regeneration attacks, proposed by Zhao et al., attempt to remove watermarks by adding Gaussian noise to the latent space embedding of a watermarked image, then generate a new image given this modified embedding \citep{regeneration-attack}. The attack is a tradeoff between resulting image quality and attack effectiveness. This attack is highly effective against Stable Signature and StegaStamp, but the authors note that their attacks were not as successful on semantic watermarking methods such as Tree-Ring. This is extended in the WAVES benchmark paper~\citep{WAVES} to rinsing attacks by running multiple regeneration attacks on an image.

\subsection{Diffusion Models, Denoising Diffusion Probabilistic Models, and Denoising Diffusion Implicit Models}

Diffusion models generate images by iteratively converting random noise into an image distribution \citep{DM}, with models like DDPMs and DDIMs enhancing stability and efficiency \citep{DDPM, DDIM}. Latent diffusion models operate in a latent space, then map to image space using a variational autoencoder \citep{stable-diffusion}.

\subsection{GradCAM - Gradient weighted Class Activation Mapping}

Decisions made by Convolutional Neural Networks can be traced back to the input via the use of GradCAM to produce a localization map that highlights the regions of interest that the network used to produce the final output. Presented in Selvaraju et al.~\citep{grad-cam}, this approach can be applied to a variety of CNN models and families to explain their final outputs based on highlighted regions in the input via the backwards flow of gradients from the output layer. For example, in a classification scenario, gradients for 'dog' would flow back to the region of the image where a dog would be depicted.

% \subsection{Watermarking methods for other modalities}

% Watermarking in other modalities, such as text and audio generative outputs, 

% Invisible Watermarking for Audio Generation Diffusion Models (Cao et al): Given an audio watermark trigger, alter the Gaussian distribution of the diffusion model to a new watermarking distribution based on the trigger.

% A Watermark for Large Language Models: Promote use of a generated set of "green" words during text generation. Uses the amount of "green" words in any given piece of text to determine if it is statistically likely for the text to be generated. 

\section{Baseline Models}
For our watermarking baseline, we use StegaStamp and Tree-Ring. StegaStamp is selected for its robustness to recent attacks, such as regeneration and adversarial attacks, despite being older than techniques like Tree-Ring and Stable Signature. Tree-Ring, a more recent method, is included as it is a standard benchmark (WAVES~\citep{WAVES}) and operates in a different domain (modifying latent space vs. pixel space for StegaStamp). WAVES highlights Tree-Ring's robustness to attacks like image blurring, which may complement StegaStamp when integrated.

For our watermark attacking baseline, we use Rotation Distortion (75°), Blur Distortion (8x8 Gaussian kernel), and Regeneration Attacks. Rotation and blurring are chosen for significantly reducing the performance of Tree-Ring and StegaStamp, albeit with reduced image quality (WAVES~\citep{WAVES}). Regeneration is included for its strong overall degradation across watermarks while maintaining better image quality, making it a solid comparison for our proposed attack.

% Dongjun Hwang
\subsection{Baseline Model Descriptions}
\label{sec:baseline model descriptions}

\begin{wraptable}{r}{0.38\textwidth}
\vspace{-1.2em}
\centering
\small
\setlength{\tabcolsep}{0.25em}
\renewcommand{\arraystretch}{1.1}
\caption{\small
    \textbf{The evaluation results of StegaStamp and our reproduced model.}}
% \vspace{1.em}
\begin{tabular}{@{}lcc@{}}
\toprule
 & StegaStamp & Ours \\ \midrule \midrule
Bit Accuracy (\%) & 0.999 & 0.997 \\ \bottomrule
\end{tabular}
\vspace{-.5em}
\label{tab:stegastamp_reproducing}
\end{wraptable}

\subsubsection{StegaStamp}
The StegaStamp encoder~\cite{stegastamp} embeds a bitstring within an image, creating an output that appears visually unchanged. It processes an input image $\mathbf{I}$ and a bitstring $\mathbf{M}$, transforming the bitstring into a tensor and combining it with the image via an encoder network to produce a residual tensor $\mathbf{R}$. The encoded image is then $\mathbf{I}_{\text{enc}} = \mathbf{I} + \mathbf{R}$, where $\mathbf{R}$ is designed to differ minimally from the original image. The decoder retrieves the bitstring from the encoded image, correcting distortions with a Spatial Transformer Network and reconstructing the bitstring via a decoder network.

The decoder is trained with the following loss function to balance regularization, perceptual similarity, and recovery accuracy

\begin{equation}
L = \lambda_R L_R + \lambda_P L_P + \lambda_M L_M
\end{equation}

where $ L_R $ is the $ L_2 $ regularization loss, $ L_P $ is the perceptual similarity loss (LPIPS), and $ L_M $ is the cross-entropy loss for bitstring recovery.

The decoder is trained with a cross-entropy loss to minimize the discrepancy between the reconstructed bitstring $ \mathbf{\hat{M}} $ and the actual bitstring $ \mathbf{M} $:

\begin{equation}
L_M = - \sum_{i=1}^{k} \left( M_i \log(\hat{M}_i) + (1 - M_i) \log(1 - \hat{M}_i) \right)
\end{equation}

For reproduction, StegaStamp was trained on the MSCOCO 2017 train set, using 400x400 resampled images and random binary messages. Evaluation with the message "Stega!!" confirms performance consistency with the original, as shown in Table~\ref{tab:stegastamp_reproducing}.

\subsubsection{Tree-Ring}

\begin{table*}[ht]
\centering
\small
\begin{minipage}{0.48\textwidth}
    \centering
    \setlength{\tabcolsep}{0.35em}
    \renewcommand{\arraystretch}{1.1}
    \captionof{table}{\small \textbf{Unattacked evaluation results of Tree-Ring and our reproduced model.}}
    \begin{tabular}{lccc} 
      \hline
      Model & AUC & TPR@1\%FPR & FID \\ 
      \hline \hline
      Original Tree-Ring & 1.000 & 1.000 & 25.93 \\ 
      Our Tree-Ring & 1.000 & 0.996 & 24.63 \\ 
      \hline
    \end{tabular}
    \label{tab:tree-ring-unattacked-reproduction}
\end{minipage}
\hfill
\begin{minipage}{0.48\textwidth}
    \centering
    \setlength{\tabcolsep}{1.2em}
    \renewcommand{\arraystretch}{1.1}
    \captionof{table}{\small \textbf{Attacked evaluation results of Tree-Ring and our reproduced model with AUC metric.}}
    \begin{tabular}{lcc} 
      \hline
      Model & Rotation & Blur \\ 
      \hline \hline
      Original Tree-Ring & 0.935 & 0.999\\
      Our Tree-Ring & 0.463 & 0.965 \\ 
      \hline
    \end{tabular}
    \label{tab:tree-ring-attacked-reproduction}
\end{minipage}

\end{table*}

In Tree-Ring~\citep{tree-ring}, the diffusion model $\epsilon_\theta$ generates images from an initial noise vector $\textbf{x}_T$ and retrieves approximations of the initial noise from an image. The watermark is embedded by injecting a circular key $k$ into the Fourier space of the noise vector before image generation. To decode, the image is renoised, and the Fourier transform is applied to compare the latent values to the original key. If the $l2$ distance between the latent values and the original key is sufficiently small, the watermark is detected.

% \begin{algorithm}
% \caption{Tree-Ring$_{rings}$ Watermarking}
% \begin{algorithmic}[1]
% \Require Initial noise vector $\bold{x}_T$ and diffusion model $\epsilon_\theta$.
% \Ensure Watermarked image $\bold{x}_0$ with radius $r$ $rings$ watermark in channel $c$.
% \State $k \in \mathbb{C}^r \sim \mathcal{N}(0, I_d)$ \Comment{Sample key values}
% \State $\bold{x}_T' = \mathcal{F}(\bold{x}_T)$ \Comment{Apply Fourier Transform}
% \State $\forall i (1\leq i \leq r), \forall (x,y) ((i-1)^2 < x^2 + y^2 \leq i^2), \bold{x}_T'[c, x, y]=k[i] $ \Comment{Inject key}
% \State $\bold{x}_T'' = \mathcal{F}^{-1}(\bold{x}_T')$ \Comment{Apply Inverse Fourier Transform}
% \State \Return $\epsilon_\theta(\bold{x}_T'')$
% \end{algorithmic}
% \label{alg:tree-ring-application}
% \end{algorithm}

% \begin{algorithm}
% \caption{Tree-Ring$_{rings}$ Watermark Detection}
% \begin{algorithmic}[1]
% \Require Image $\bold{x}_0$, key $k$, key mask $M$, and diffusion model $\epsilon_\theta$.
% \Ensure Detection probability $p$.
% \State $\bold{x}_T = \epsilon_\theta(\bold{x}_0)$ \Comment{Get approximate initial noise vector}
% \State $\bold{x}_T' = \mathcal{F}(\bold{x}_T)$ \Comment{Apply Fourier Transform}
% \State $\sigma^2=\frac{1}{|M|}\sum_{i\in M}|y_i^2|$ \Comment{Get estimated variance}
% \State $\eta = \frac{1}{\sigma^2}\sum_{i\in M}|k_i - y_i|^2$ \Comment{Calculate score}
% \State $p=\text{Pr}\big(\chi^2_{|M|, \lambda}\leq \eta \big| H_0\big)=\Phi_{\chi^2}(\eta)$ \Comment{Get p-value using non-central $\chi^2$ CDF}
% \State \Return $p$
% \end{algorithmic}
% \label{alg:tree-ring-detection}
% \end{algorithm}

In our reimplementation of the code in section 3.3.2, we use the same Tree-Ring radius of 10 and guidance scale of 7.5 as the original paper. As seen in tables~\ref{tab:tree-ring-unattacked-reproduction} and~\ref{tab:tree-ring-attacked-reproduction}, our reproduction achieves very similar unattacked performance and blurred attack performance on the 5000 unwatermarked and 5000 watermarked images generated from the MSCOCO 2017 validation set captions. 

However, our model performs much worse against rotation attacks than the original paper. We think that this is because a rotation in the pixels space does not necessarily correspond to a rotation in the latent space, which leads to the Fourier space latents not looking like the original key. This effect is demonstrated in figure~\ref{fig:fourier-renoised-latents-tree-ring} of the appendix. This issue persists through variations in the placement of the watermark in different channels, switching between Stable Diffusion 2.1 base and Stable Diffusion 2, and switching schedulers from DPMSolverMultistepScheduler and DDIMScheduler. This demonstrates a weakness of the Tree-Ring model -- it is highly dependent on the renoised latents being similar to the original latents in order to get a strong detection.

\subsubsection{Regeneration and Rinsing Attacks}

Regeneration attacks perturb an image's latent representation by adding noise to its latent space and then denoising it~\citep{regeneration-attack}. Let $\phi: \mathbb{R}^n \rightarrow \mathbb{R}^d$ be an embedding function returning an image's representation, $\mathcal{A}: \mathbb{R}^d \rightarrow \mathbb{R}^n$ a regeneration function reconstructing an image from its embedding, and $x_w \in \mathbb{R}^n$ a watermarked image. A single regeneration attack produces an image $\hat{x}$ as follows:

$$\hat{x} = \mathcal{A}( \phi(x_w) + \mathcal{N}(0, \sigma^2 I_d))$$

Here, $\phi(x_w)$ represents the watermarked image's embedding, which is perturbed by Gaussian noise $\mathcal{N}(0, \sigma^2 I_d)$ and then passed through $\mathcal{A}$ to produce a clean output image. Zhao et al.~\citep{regeneration-attack} tested various embedding and regeneration method combinations, finding that the Diffusion model regeneration attack was consistently effective across watermarking schemes. In this method, $\phi$ and $\mathcal{A}$ use pretrained Stable Diffusion models.

The WAVES Benchmark paper~\citep{WAVES} introduces Rinsing attacks, which apply multiple iterations of the Regeneration attack. The output of each iteration serves as input for the next. The performance of these attacks depends on the iteration count and the noise added per iteration. The WAVES paper suggests that two iterations with a timestep of 20-100 per diffusion strike a balance between low TPR@0.1\%FPR and high image quality.

% We evaluated the baseline implementation on 5000 Stable Signature watermarked images from the Erasing the Invisible watermark competition warm-up kit. Although the original Regeneration attack paper~\citep{regeneration-attack} did not assess this method on Stable Signature watermarks, the WAVES Benchmark paper found Regeneration attacks ranked highest or near-highest among evaluated methods~\citep{WAVES}, effectively destroying the watermark with minimal impact on image quality. While AUC and FID metrics were not reported, their average TPR@0.1\%FPR is summarized in table~\ref{fig:regen-rinse-table}.

\begin{table}[ht]
\begin{center}
\small
\setlength{\tabcolsep}{1.2em}
\renewcommand{\arraystretch}{1.1}
\captionof{table}{\small \textbf{Evaluation results of single regeneration attack against Stable Signature watermarked images.} The attacks are denoted as [iterations]x[strength], where strength is the timestep parameter. The resulting TPR@0.1\%FPR, AUC, and FID against unwatermarked images are provided. *Note that the WAVES Benchmark does not directly provide AUC/FID information for us to compare against, and thus are listed as N/A. } 
\vspace{1.em}
\begin{tabular}{lcccc} 
  \hline
  Model & TPR@0.1\%FPR & AUC & FID against original images\\ 
  \hline \hline
  Unattacked (FID) & N/A & N/A & 24.93\\ 
  Regeneration Attack, 1x60 & 0.000 & 0.562 & 23.57 \\ 
  Rinsing Attack, SD v1.4 2x20 & 0.000 & 0.564 & 23.51 \\
  Rinsing Attack, SD v2.1 2x10 & 0.000 & 0.564 & 23.61 \\
  WAVES Regeneration Benchmark & 0.000 & N/A* & N/A* \\
  WAVES Rinsing Attack & 0.000 & N/A* & N/A* \\
  \hline
\end{tabular}
\label{fig:regen-rinse-table}
\end{center}
\end{table}

In our reproduction, we implemented Regeneration and Rinsing attacks using Stable Diffusion v1.4 checkpoints for comparison to the WAVES paper~\citep{WAVES}. We also included a Rinsing attack using Stable Diffusion v2.1 checkpoints for additional comparison. We evaluated the baseline implementation on 5000 Stable Signature watermarked images from the Erasing the Invisible watermark competition warm-up kit~\footnote{\url{https://github.com/erasinginvisible/warm-up-kit/tree/main}}. Performance was assessed by TPR@0.1\%FPR, AUC, and FID on the Stable Signature watermarked images. All three attacks excelled against Stable Signature watermarks, as shown in Table~\ref{fig:regen-rinse-table}. Their FID values indicated minimal quality degradation compared to unattacked images. These results confirm that diffusion-based Regeneration and Rinsing attacks effectively remove Stable Signature watermarks with negligible impact on image quality, aligning with WAVES Benchmark findings~\citep{WAVES}.

\section{Proposed Models}
In our project, we aim to improve the efficacy of both watermarking methods and watermark attack methods. We divide our project scope into two sections below, where the first part explores improving watermark robustness by using multiple watermarking techniques in tandem, and the second part explores localizing attacks to target regions of the target image to achieve lower image degradation.

\subsection{Proposed Robust Watermarking Methods}

\subsubsection{Naive Stacking Model}

\begin{figure}[ht]
    \centering
    \includegraphics[width=0.98\textwidth]{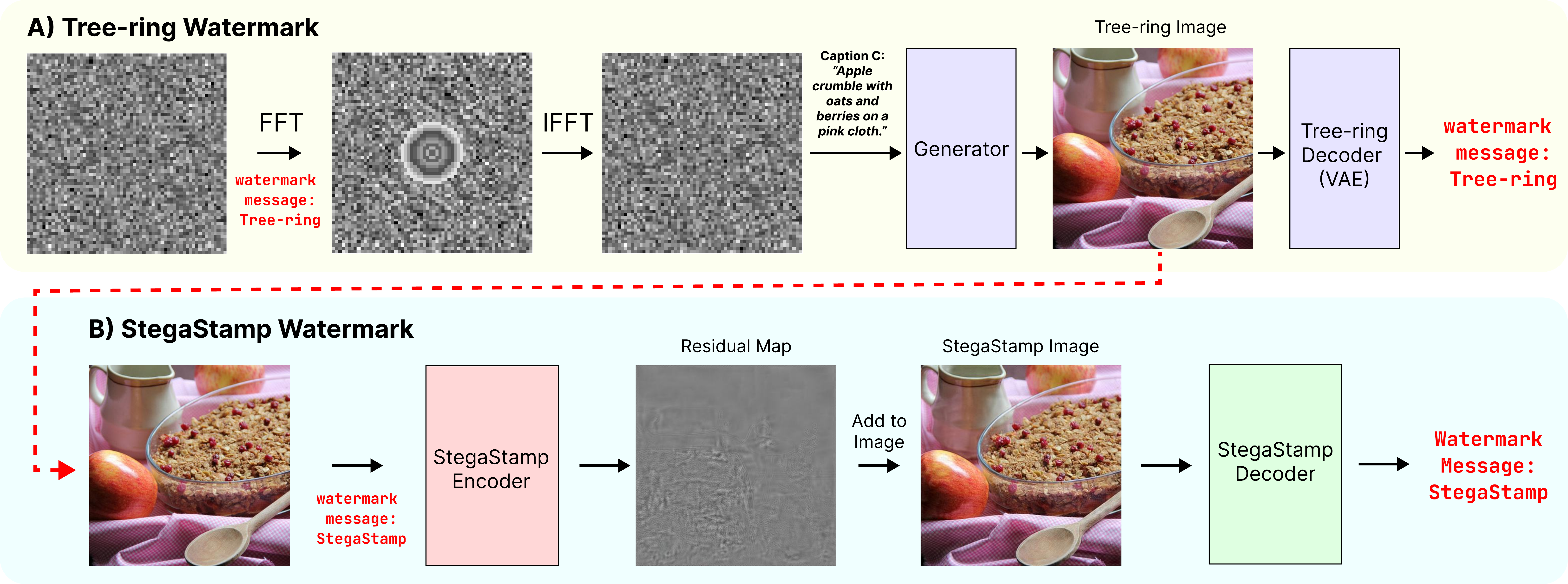}
    \caption{\small
    \textbf{Overview of the naive stacking of Tree-ring and StegaStamp watermarking pipeline.}}
    \label{fig:stacking_watermark}
\end{figure}

A straightforward approach to creating a robust watermark involves stacking two techniques. As shown in Figure~\ref{fig:stacking_watermark}, we combine Tree-Ring and StegaStamp by sequentially applying both. Tree-Ring embeds a watermark during image generation, so it must precede StegaStamp, which works on pre-generated images.

\begin{wraptable}{r}{0.45\textwidth}
\vspace{-1.2em}
\setlength{\tabcolsep}{0.8em}
\renewcommand{\arraystretch}{1.}
\centering
\small
\caption{Average $\ell_2$ distance between corresponding watermarked and non-watermarked images for each method.}
\vspace{-1.em}
\begin{tabular}{lcc}\\\toprule  
Method & \begin{tabular}{@{}c@{}}Image \\ $\ell_2$ distance\end{tabular} & \begin{tabular}{@{}c@{}}Latent \\ $\ell_2$ distance\end{tabular} \\ \midrule \midrule
StegaStamp  & 17.40 & 118.17 \\  
TreeRing & 117.58 & 52.81\\  \bottomrule
\end{tabular}
\vspace{-1.em}
\label{tab:watermark_l2_dist}
\end{wraptable} 

However, applying these methods sequentially is challenging due to their impact on different feature spaces. Referenced from Saberi et al.~\citep{saberi2023robustness}, table~\ref{tab:watermark_l2_dist} shows that Tree-Ring alters pixel space, while StegaStamp operates in latent space. Applying StegaStamp on top of Tree-Ring would disturb the latter's watermark, making it undetectable. Thus, a direct stacking approach is impractical, and we propose a remover architecture to address this.

\subsubsection{Remover Architecture Stacking Model}

\begin{figure}[ht]
    \centering
    \includegraphics[width=0.98\textwidth]{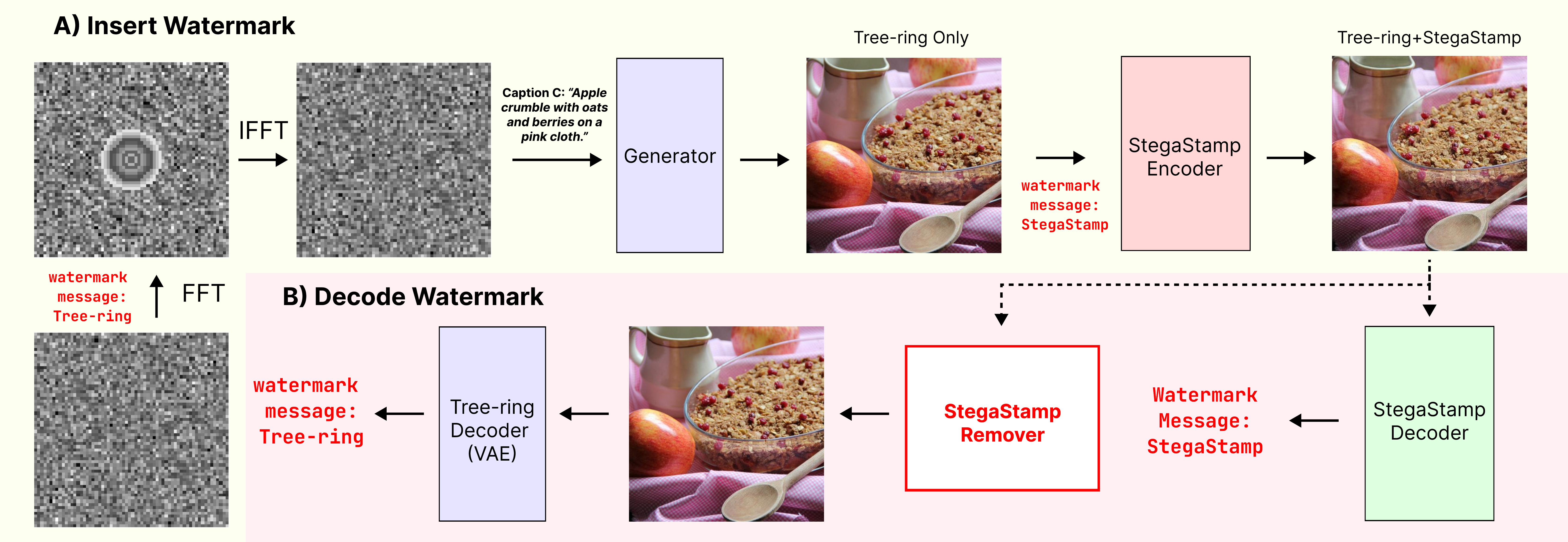}
    \caption{\small
    \textbf{Overview of the stacking of the modified Tree-ring and StegaStamp watermarking pipeline with the remover network.}}
    \label{fig:remover}
\end{figure}

To resolve this, we introduce a dedicated network to remove the StegaStamp watermark, reducing perturbations from overlapping watermarks. The process is shown in Figure~\ref{fig:remover} and detailed in Algorithm~\ref{alg:remover}.

\begin{algorithm}
\small
\caption{Stacking Tree-ring and StegaStamp Watermarking Pipeline}
\begin{algorithmic}[1]
\State \textbf{Input:} Original image $I$, Tree-ring message $M_{TR}$, StegaStamp message $M_{SS}$
\State \textbf{Output:} Watermarked image $I_{TR+SS}$, Decoded messages $M_{TR}, M_{SS}$

\State \textbf{Step 1: Tree-ring Watermark Insertion}
\State Compute the FFT of the image: $I_{FFT} \gets \text{FFT}(I)$
\State Embed the Tree-ring message: $I_{TR\_FFT} \gets \text{Embed}(I_{FFT}, M_{TR})$
\State Compute the IFFT to return to the spatial domain: $I_{TR} \gets \text{IFFT}(I_{TR\_FFT})$

\State \textbf{Step 2: StegaStamp Watermark Insertion}
\State Pass the Tree-ring watermarked image through the StegaStamp encoder: $I_{TR+SS} \gets \text{Encoder}_{SS}(I_{TR}, M_{SS})$

\State \textbf{Step 3: StegaStamp Watermark Decoding}
\State Pass $I_{TR+SS}$ through the StegaStamp decoder: $M_{SS} \gets \text{Decoder}_{SS}(I_{TR+SS})$

\State \textbf{Step 4: StegaStamp Watermark Removal}
\State Remove the StegaStamp watermark: $I_{TR\_Removed} \gets \text{Remover}_{SS}(I_{TR+SS})$

\State \textbf{Step 5: Tree-ring Watermark Decoding}
\State Compute the FFT of $I_{TR\_Removed}$: $I_{TR\_Removed\_FFT} \gets \text{FFT}(I_{TR\_Removed})$
\State Decode the Tree-ring message: $M_{TR} \gets \text{Decoder}_{TR}(I_{TR\_Removed\_FFT})$

\end{algorithmic}
\label{alg:remover}
\end{algorithm}

\begin{figure}[ht]
    \centering
    \begin{minipage}{0.46\textwidth}
        \centering
        \includegraphics[width=\textwidth]{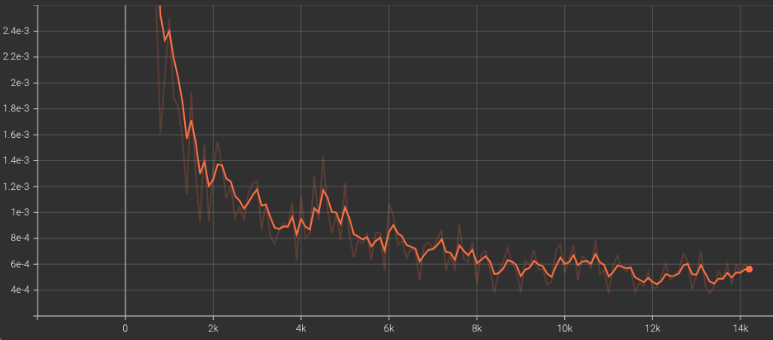}
        \caption{\small
        \textbf{Loss plot when training the stegastamp remover network.}}
        \label{fig:remover_loss}
    \end{minipage}%
    \hfill
    \begin{minipage}{0.5\textwidth}
        \small
        \centering
        \captionof{table}{\small
        \textbf{Final training loss and validation loss.}}
        \vspace{1.em}
        \begin{tabular}{@{}lcc@{}}
        \toprule
         & Training Loss & Validation Loss \\ \midrule \midrule
        Remover Network & 0.00057 & 0.00108 \\ \bottomrule
        \end{tabular}
        \label{tab:remover_loss_table}
    \end{minipage}
\end{figure}

We train the Remover Network to minimize the perturbation caused by overlapping watermarks during decoding. The network minimizes the L2 distance between the output image, \(I_{\text{TR\_Removed}}\), and the original Tree-ring watermarked image, \(I_{\text{TR}}\). The loss function is defined as:

\begin{equation}
\mathcal{L}_{\text{remover}} = \frac{1}{N} \sum_{i=1}^{N} \| I_{\text{TR\_Removed}}^{(i)} - I_{\text{TR}}^{(i)} \|_2^2,
\end{equation}

where \(N\) denotes the number of samples in the batch. By optimizing this objective, the network learns to effectively remove the StegaStamp watermark from \(I_{\text{TR+SS}}\), producing an output \(I_{\text{TR\_Removed}}\) that closely approximates \(I_{\text{TR}}\). As shown in Figure~\ref{fig:remover_loss} and Table~\ref{tab:remover_loss_table}, both training and validation losses converge well.

After training, the Remover Network can decode the stacked watermarks. The complete pipeline is shown in Figure~\ref{fig:remover} and Algorithm~\ref{alg:remover}, detailing watermark embedding, decoding, and interference reduction.

\subsection{Proposed Watermark Attack Method: Localized Blurring Attack (LBA)}

The attacks listed in the WAVES Benchmark~\citep{WAVES} target the entire image-space or latent-space. However, a uniform attack across the whole image is unnecessary and can degrade perceptual quality. In this work, we focus on StegaStamp, which embeds information using steganography in the image-space. We investigate a localized blurring attack (LBA), as it is easy to implement and effective, especially at higher strengths in the image-space.

\begin{figure}[ht]
    \centering
    \includegraphics[width=0.95\textwidth]{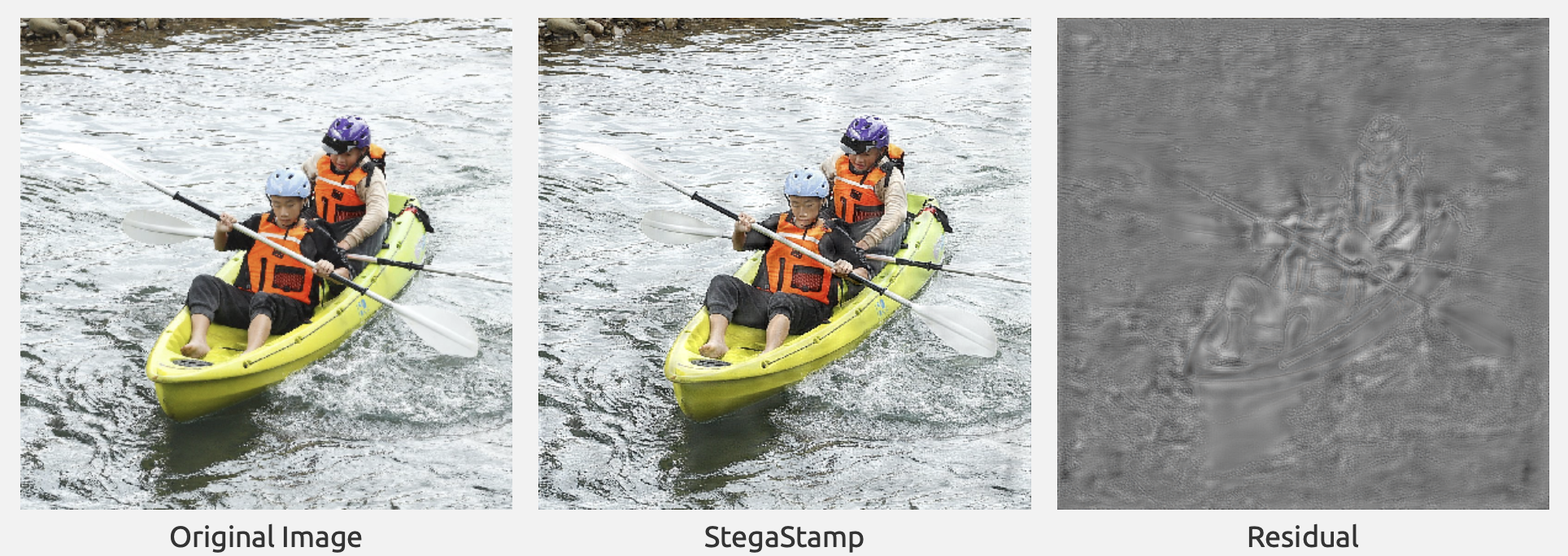}
    \caption{\small
    \textbf{Example of StegaStamp residuals on a watermarked image.} The residuals are mostly applied around the kayak and the two occupants. Image taken from~\citep{stegastamp}.}

    \label{fig:stega_residual_kayak}
\end{figure}

Figure~\ref{fig:stega_residual_kayak} shows that StegaStamp residuals (the difference between watermarked and unwatermarked images) are concentrated in visually interesting areas with higher contrast, such as the kayak and riders, rather than in flatter regions like the water. The StegaStamp watermark decoder processes the watermarked image through convolutional layers, applies a Sigmoid function, and rounds the results to return binary bits. By using GradCAM with the decoder, we visualize areas with the highest gradients.

\begin{figure}[ht]
    \centering
    \includegraphics[width=0.95\textwidth]{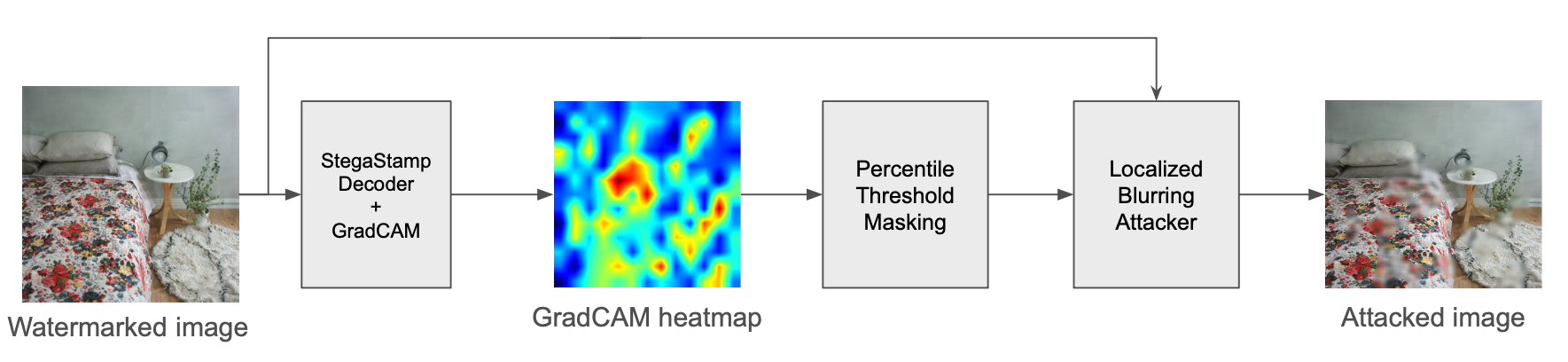}
    \caption{\small
    \textbf{Localized Blurring Attack pipeline.}}
    \label{fig:lba_pipeline}
\end{figure}

We propose the Localized Blurring Attack (LBA), which applies a blurring kernel to selected regions of the image. The process, shown in Figure~\ref{fig:lba_pipeline}, begins by running the watermarked image through the StegaStamp decoder with GradCAM to generate a heatmap. Pixels above a specified percentile are thresholded to create a binary mask, and these regions are replaced with blurred pixels from the original image. A visualization for percentile thresholding is available in Figure~\ref{fig:lba-percentile-threshold}.

Key variables for this method include the kernel size and percentile threshold, which we will evaluate in the results section.

\section{Dataset Description}

\myparagraph{MS-COCO 2017 validation split.} For the majority of our work, we evaluate the performance and quality of the resulting images based off of the 5,000 images in the validation split of Microsoft Common Objects in Context (MSCOCO)~\citep{ms-coco}. For the purposes of generating Tree-Ring image, MSCOCO also includes image captions that can be used for prompting the selected diffusion model.

% \myparagraph{NeurIPS 2024 Erasing the Invisible Challenge Warm-up Kit.}  The NeurIPS 2024 Erasing the Invisible Challenge~\footnote{\url{https://erasinginvisible.github.io/}} is a watermark attacking competition. For this competition, a warm-up kit is provided~\footnote{\url{https://github.com/erasinginvisible/warm-up-kit/tree/main}} with 5,000 paired sets of Stable Signature watermarked images and unwatermarked images. Some of our baseline attacks are evaluated with this set of images.
% \myparagraph{NeurIPS 2024 Erasing the Invisible Challenge.} The NeurIPS 2024 Erasing the Invisible Challenge~\footnote{\url{https://erasinginvisible.github.io/}} is a watermark attacking competition. There are two tracks: beige box and black box. In the beige box track, 300 watermarked images are provided, 150 of which are watermarked with (a variant) of StegaStamp, and the rest watermarked with TreeRing. In the black box track, 300 watermarked images are given without knowledge of the watermarking algorithm. These two datasets will serve as our baseline attack dataset.

% \myparagraph{Other Evaluation Datasets.} We evaluate our method using images and captions from the 2017 validation split of MS-COCO~\citep{ms-coco}. Specifically for the adversarial attack baseline, we evaluate on ImageNet~\citep{imagenet}.

\section{Metric Descriptions}

\subsection{Fréchet Inception Distance (FID)}

The Fréchet Inception Distance (FID) \cite{heusel2017gans} compares a generative model's image set to a reference dataset of real-world images by modeling both as multidimensional Gaussian distributions $\mathcal{N}(\mu, \Sigma)$. The distributions are derived by running both image sets through the Inception v3 CNN and comparing the activation vectors at its deepest layer. This comparison measures high-level feature similarities and serves as a proxy for image quality. FID can be computed using the Python library pytorch-fid.

\subsection{AUC and TPR@1\%FPR}

TPR@1\%FPR represents the true positive rate (TPR) at a 1\% false positive rate (FPR), indicating the proportion of correctly identified watermarked or unwatermarked images at this FPR. The Area Under the Curve (AUC) measures the probability that a model ranks a positive example higher than a negative one, calculated from the receiver operating characteristic (ROC) curve. A perfect model has an AUC of 1. The Python library scikit-learn can compute the ROC curve, AUC, and TPR@1\%FPR.

\subsection{Bit Accuracy and Detection Rate}

Bit accuracy evaluates the correctness of classification models by checking if each predicted bit matches the actual bit:

\begin{equation} \text{Bit Accuracy} = \frac{1}{N} \sum_{n=1}^{N} \mathbbm{1} (\hat{y}_{n} = y_{n}) \end{equation}

where $\hat{y}_{n}$ is the predicted bit and $y_{n}$ is the actual bit. The detection rate uses the same formula, but $\hat{y}_{n}$ indicates whether a watermark is detected, and $y_{n}$ shows if the image is watermarked. For StegaStamp, a watermark is detected if the decoder successfully decodes a message, and for Tree-Ring, if the p-value falls below 0.01.

\section{Results and Discussion}
\subsection{Combining StegaStamp and Tree-Ring}

\begin{table}[ht]
\centering
\caption{Comparison of Bit Accuracy for StegaStamp and AUC/TPR@1\%FPR for Tree-Ring}
\begin{minipage}[t]{0.45\linewidth}
    \centering
    \small
    \setlength{\tabcolsep}{0.3em}
    \renewcommand{\arraystretch}{1.4}
    \caption*{(a) Bit Accuracy Comparison for StegaStamp watermark}
    \begin{tabular}{lc}
    \hline
    \textbf{Method}      & \textbf{Bit Accuracy } \\ \hline \hline
    StegaStamp           & 0.997                      \\ 
    Naively Stacking     & 0.998                      \\ \hline
    \end{tabular}
    \label{table:result_ss}
\end{minipage}%
\hfill
\begin{minipage}[t]{0.52\linewidth}
    \centering
    \small
    \setlength{\tabcolsep}{0.3em}
    \renewcommand{\arraystretch}{1.4}
    \caption*{(b) AUC and TPR@1\%FPR Comparison for Tree-Ring watermark}
    \begin{tabular}{lccc}
    \hline
    \textbf{Metric}     & \textbf{Tree-Ring} & \textbf{+ Stacked} & \textbf{+ Remover} \\ \hline \hline
    AUC                 & 0.9956                     & 0.9936                     & 0.9950                     \\ 
    TPR@1\%FPR          & 0.96                       & 0.95                       & 0.96                       \\ \hline
    \end{tabular}
\end{minipage}
\end{table}

Table~\ref{table:result_ss}(a) shows the Bit Accuracy of StegaStamp and a naively stacked approach, with StegaStamp achieving 0.997 and the naively stacked approach achieving a slightly higher value of 0.998. Both methods demonstrate near-perfect bit accuracy, indicating their effectiveness in embedding and retrieving watermarks. Notably, the naively stacked method combines multiple watermarks without interfering with their individual attributes, allowing the two watermarks to coexist and complement each other effectively. This result highlights the robustness of stacking and its potential for designing multi-layered watermarking systems.

Table~\ref{table:result_ss}(b) evaluates the AUC and TPR@1\%FPR metrics for Tree-Ring watermarks. The original Tree-Ring watermark achieves an AUC of 0.9956 and a TPR@1\%FPR of 0.96. When combined with a naive stacked approach, the performance slightly decreases to an AUC of 0.9936 and a TPR@1\%FPR of 0.95, but the overall performance remains robust, showing that the stacked watermarks do not interfere with each other. After applying a watermark remover, the AUC recovers to 0.9950, and the TPR@1\%FPR remains at 0.96, closely matching the original Tree-Ring performance. These results demonstrate that the remover effectively isolates and enhances the Tree-Ring watermark, allowing for accurate detection without degradation.

\begin{table}[ht]
\centering
\small
\setlength{\tabcolsep}{0.3em}
\renewcommand{\arraystretch}{1.4}
\caption{P-Value Comparison}
\begin{tabular}{lcccc}
\hline
Metric     & Unwatermarked & Watermarked Image & + Naively Stacking & + Remover Network \\ \hline \hline
P-Value             & 0.490                       & 0.009                      & 0.014                       & 0.010                      \\ \hline
\end{tabular}
\label{tab:stacking_bit_acc}
\end{table}

Table~\ref{tab:stacking_bit_acc} compares p-values across four scenarios: Unwatermarked images, Watermarked images, Naively Stacked watermarked images, and images processed through the Remover Network. Unwatermarked images exhibit a high p-value of 0.490, indicating low probability of watermark presence, while watermarked images have a low p-value of 0.009. Naively stacked images maintain a p-value of 0.014, showing that stacking does not interfere with detectability. After applying the Remover Network, the p-value remains low at 0.010, suggesting the remover effectively isolates the watermark signal without degrading its detectability.

\begin{table}[h!]
\centering
\small
\setlength{\tabcolsep}{0.3em}
\renewcommand{\arraystretch}{1.4}
\caption{Detection rate for different methods under various transformations.}
\begin{tabular}{lccc}
\hline
\textbf{Method} & \textbf{None} & \textbf{Blurring} & \textbf{Rotation} \\
\hline \hline
Original Tree-ring & 0.911 & 0.218 & 0.000 \\
Original StegaStamp & 1.000 & 0.089 & 0.010 \\
Naively Stacking & 1.000 & 0.158 & 0.000 \\
Remover & 1.000 & 0.168 & 0.000 \\
\hline
\end{tabular}
\vspace{-1.em}
\label{table:detection_rate}
\end{table}

Table~\ref{table:detection_rate} evaluates the detection rate of StegaStamp and Tree-Ring under various transformations, including Blurring and Rotation. Without any attack (None), our method achieves a detection rate of 1.000, outperforming the original Tree-Ring’s 0.911, highlighting improved detection capabilities from StegaStamp. Under Blurring, the detection rate increases after passing through the remover (0.168 for remover vs. 0.158 for naively stacked), showing that the remover effectively detects Tree-Ring while mitigating StegaStamp interference. Furthermore, it also shows that combining the two improves on StegaStamp's robustness to blurring. However, the combined watermarks perform worse than Tree-Ring on its own for blurring, suggesting that even after the remover is applied there is some latent space interference from StegaStamp. Furthermore, all methods fail to detect watermarks under Rotation, revealing a limitation in robustness against rotational transformations. This indicates that combining two watermarks does not bring extra robustness to attacks they do not perform well against, highlighting the need for further improvements to enhance resilience against such attacks.

\subsection{Localized Blurring Attack}

\begin{table}[h!]
\vspace{-1.em}
\centering
\caption{Detection Rates and FIDs for Different Attacks and Configurations}
\begin{adjustbox}{width=\textwidth}
\begin{tabular}{@{}cccccccccc@{}}
\toprule
\multirow{2}{*}{Attack Type} & \multirow{2}{*}{Percentile} & \multicolumn{3}{c}{Blur Kernel Size (Detection Rates)} & \multicolumn{3}{c}{Blur Kernel Size (FID)} & \multicolumn{2}{c}{Regeneration Strength 60} \\
\cmidrule(lr){3-5} \cmidrule(lr){6-8} \cmidrule(lr){9-10}
 &  & 5 & 11 & 31 & 5 & 11 & 31 & Detection Rate & FID \\
\midrule
\multirow{4}{*}{Localized Blurring (LBA)} & 0 & 1.0000 & 0.2594 & 0.0366 & 21.9822 & 50.1141 & 118.6756 & - & - \\
 & 25 & 1.0000 & 0.4908 & 0.0376 & 21.6435 & 54.2929 & 136.1907 & - & - \\
 & 50 & 1.0000 & 0.8536 & 0.0554 & 18.8150 & 40.4577 & 88.0998 & - & - \\
 & 75 & 1.0000 & 0.9980 & 0.5372 & 12.1990 & 20.4747 & 36.0739 & - & - \\
\midrule
\multirow{4}{*}{Randomized Attack} & 0 & 1.0000 & 1.0000 & 0.9874 & 7.8892 & 11.4521 & 17.2926 & - & - \\
 & 25 & 1.0000 & 1.0000 & 0.9882 & 8.5146 & 13.2648 & 20.3889 & - & - \\
 & 50 & 1.0000 & 1.0000 & 0.9796 & 9.6937 & 16.6092 & 25.4361 & - & - \\
 & 75 & 1.0000 & 1.0000 & 0.8936 & 11.3115 & 21.3751 & 33.2434 & - & - \\
\midrule
% \multirow{4}{*}{Residual Attack} & 0 & 1.0000 & 0.1436 & 0.0000 & 20.7599 & 48.7613 & 178.5916 & - & - \\
%  & 25 & 0.9898 & 0.2322 & 0.0010 & 17.5559 & 52.2570 & 137.2971 & - & - \\
%  & 50 & 0.9146 & 0.4114 & 0.0024 & 15.8805 & 40.5224 & 82.6640 & - & - \\
%  & 75 & 0.8980 & 0.7230 & 0.1896 & 12.8534 & 25.4160 & 40.1597 & - & - \\
% \midrule
% \multirow{4}{*}{Residual Attack (Diff Msg)} & 0 & 1.0000 & 0.2660 & 0.0008 & 21.3934 & 49.1636 & 174.9596 & - & - \\
%  & 25 & 1.0000 & 0.8272 & 0.0032 & 20.0985 & 56.1382 & 142.4076 & - & - \\
%  & 50 & 0.9998 & 0.9824 & 0.0800 & 19.0279 & 45.3853 & 88.8295 & - & - \\
%  & 75 & 1.0000 & 0.9992 & 0.9710 & 14.5098 & 28.9673 & 43.2599 & - & - \\
% \midrule
Straight Blurring & - & 1.0000 & 0.1014 & 0.0000 & 26.97 & 62.39 & 150.24 & - & - \\
\midrule
Baseline Regeneration & - & - & - & - & - & - & - & 0.0100 & 16.3 \\
\bottomrule
\end{tabular}
\end{adjustbox}
% \caption{Detection rates and FID after each attack. Residual and Residual with Different Message attack the pixels from the residuals between the original image and the watermarked image. Localized Blurring Attack (LBA) attacks the GradCAM map pixels. Randomized attacks random pixels, where the proportion of random pixels picked is standardized to be the same as LBA. Blur is the baseline where all pixels are blurred. Regeneration is the baseline method mentioned earlier.}
\label{tab:LBA results}
\end{table}

Table~\ref{tab:LBA results} shows that LBA effectively reduces the detection rate of the StegaStamp watermark while balancing image quality and robustness. Increasing the percentile threshold improves FID but raises detection rates, while larger blurring kernels lower detection rates but degrade image quality. For example, with a 50\% threshold and kernel size of 31, LBA achieves a detection rate of 0.0554 and improves FID to 88.1, compared to straight blurring, which achieves a detection rate of 0.0000 but with a much higher FID of 150.24. Using a 75\% threshold further improves FID to 36.1 but raises the detection rate to 0.5372.

Compared to straight blurring, LBA maintains lower detection rates with significantly better FID, demonstrating that it reduces image degradation while maintaining performance. Additionally, LBA targets meaningful pixels, as shown by comparison with randomized attacks, which minimally affect detectability. This confirms that LBA’s improvements result from targeting key regions rather than merely attacking more pixels.

However, LBA does not outperform the baseline regeneration attack, which achieves similar detection rates with a significantly better FID of 16.3. The best FID achieved by LBA with a meaningful decrease in detectability was 36.1. This limitation likely arises from blurring's inherently destructive nature, as it does not preserve similarity to the original image and heavily damages the areas it attacks. Although LBA focuses on fewer regions, the attacked areas are completely destroyed. Furthermore, LBA assumes access to the exact StegaStamp decoder. If the attacker does not have access to the exact decoder, they may need to retrain or approximate their own, potentially leading to different results. However, in cases where an attacker can successfully reproduce a similar decoder, LBA is a practical way of decreasing the image quality degradation of attacks that they use.

\section{Future Work}

\subsection{Combining StegaStamp and Tree-Ring}
Our analysis shows that combining watermarks does not interfere with each other, challenging our initial assumption and warranting further investigation. Expanding tests of the remover network to include attacks like regeneration, beyond blurring and rotation, is crucial to assess overall robustness. For the remover network, additional training loss terms may be introduced to reduce latent space differences after removal, which may reduce degradation to the Tree-Ring watermark. Future work will also explore stacking diverse watermarking techniques beyond Tree-Ring and StegaStamp to enhance adaptability and applicability.

\subsection{Localized Blurring Attacks}
LBA assumes access to the exact StegaStamp decoder for generating the GradCAM map, which may not be realistic in all scenarios. Further studies are needed to assess these variations in cases where an attack has to retrain a new decoder that is meant to match the original decoder. Furthermore, to address LBA performing worse than regeneration, future work could improve LBA by combining it with less destructive methods like regeneration to balance image quality and degradation. We would expect localized attacks to further improve on the low image degradation shown by regeneration attacks.

\section{Conclusion}

In this paper, we explore how to combine the StegaStamp and Tree-Ring watermarks while reducing image degradation caused by localized attacks.

When combining watermarks, we examine methods for enhancing robustness by stacking different watermarks, such as Tree-Ring and StegaStamp. We find that their perturbations occur in distinct feature spaces, which can degrade detection if stacked naively. To address this, we propose a new pipeline with a remover network to prevent interference during decoding. Our experiments show that stacking watermarks alone doesn’t significantly improve recognition, but the remover network enables more effective watermarking. While stacking improves robustness compared to a single watermark, the performance gain from the remover network is smaller than expected. Future work will focus on evaluating its effectiveness against a broader range of attacks.

To reduce image degradation, we demonstrate that selectively attacking important pixels can lessen the impact of an attack. Using localized blurring attacks (LBA) with GradCAM, we identified key pixels for the StegaStamp decoder and blurred only those. This approach halved the FID (indicating better image quality) while maintaining similar performance degradation to regular blurring attacks. This has important implications for watermark design. Open-source decoders must ensure privacy to prevent attackers from using localized attacks to preserve more of the original image. Additionally, unevenly dispersed information may make watermarks vulnerable to such targeted attacks. Future watermarks should consider their concentration to mitigate this risk.

\section{Division of work}

The contributions primarily reflect the implementation work contributed by each member, each member contributed equally to ideation. Raymond contributed by reproducing Tree-Ring, implementing localized blurring attacks, helping write the evaluation code for the remover architecture, and the maintenance of the GitHub repository. Tom reproduced the regeneration attack and implemented GradCAM on the StegaStamp decoder architecture. Dongjun reproduced StegaStamp, helped implement the Remover architecture, and helped evaluated the Remover architecture. Sungwon reproduced adversarial surrogate attacks (removed from the final paper), helped implement the Remover architecture, and helped evaluated the Remover architecture.

\bibliographystyle{unsrt}

\appendix
\section{Appendix: Additional Figures}

\begin{figure}[ht]
    \begin{subfigure}{\textwidth}
        \caption{Original Watermark}
        \includegraphics[width=\linewidth]{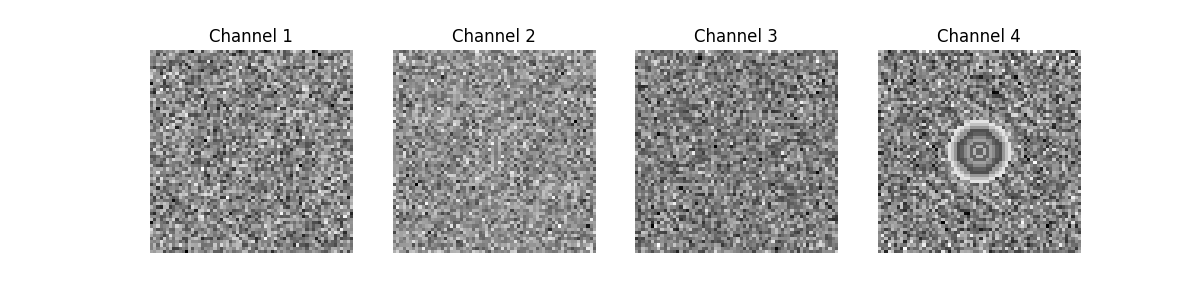}
    \end{subfigure}
    \begin{subfigure}{\textwidth}
        \caption{Renoised Unrotated Watermark}
        \includegraphics[width=\linewidth]{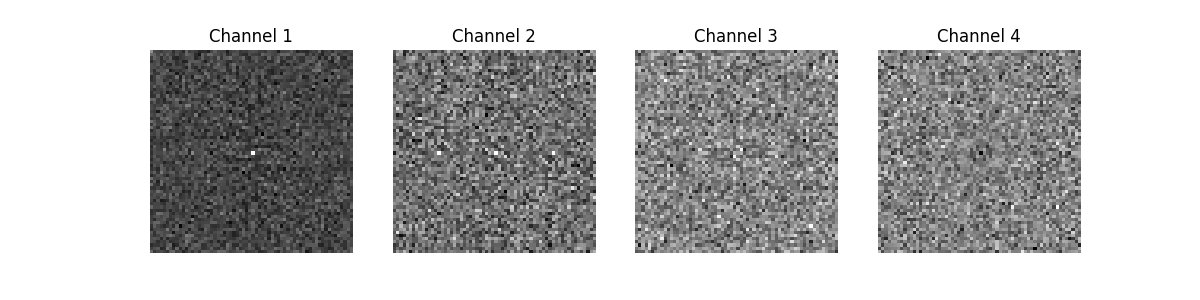}
    \end{subfigure}
    \begin{subfigure}{\textwidth}
        \caption{Renoised Rotated Watermark}
        \includegraphics[width=\linewidth]{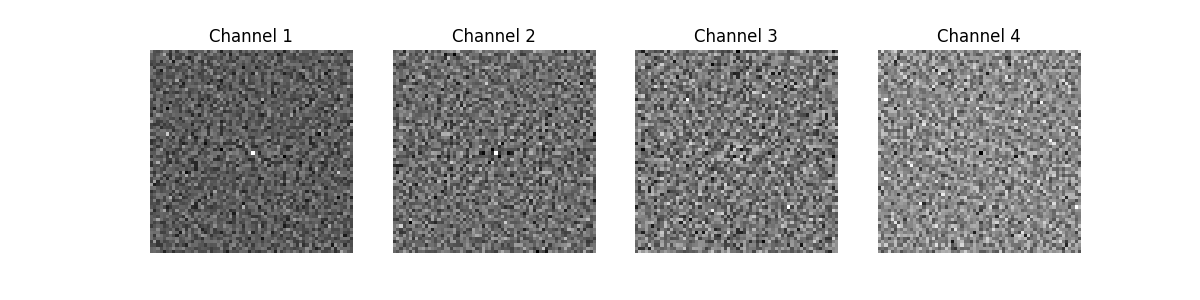}        
    \end{subfigure}
    \caption{\small\textbf{From top to bottom, the original watermarked Fourier space latents, renoised unattacked Fourier space latents, and renoised rotation attack Fourier space latents.} We can see in (b) that there is still a bit of a ring-like pattern in the center of channel 4 prior to being attacked. This is enough to be detected with a p-value of 1e-6. In subfigure (c), we can see that there is no longer any ring pattern, which suggests that rotating an image does not necessarily correspond to a rotation in the latent space (which would correspond to a rotation in the Fourier space of the latent space).}
    \label{fig:fourier-renoised-latents-tree-ring}
\end{figure}

\begin{figure}
    \centering
    \includegraphics[width=\linewidth]{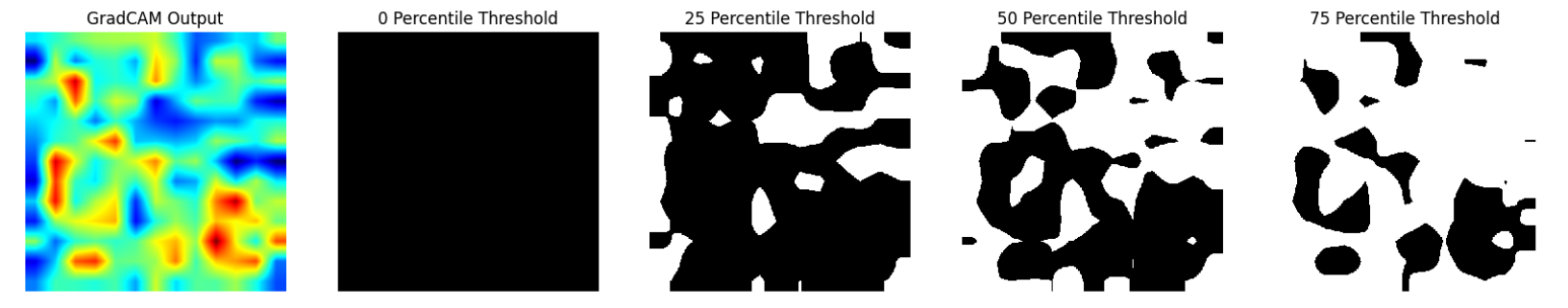}
    \caption{\small\textbf{Example binary masking from GradCAM outputs using Percentile Thresholding.} As the percentile threshold increases, smaller regions of the image are selected as the target for localized blurring attack.}
    \label{fig:lba-percentile-threshold}
\end{figure}

\end{document}